\documentclass[preprint,12pt]{elsarticle}

\usepackage{color}

\begin{document}

\begin{frontmatter}

\title{{\small Comment} \\ \vspace{0.5cm} Towards a theory of word order. Comment on ``Dependency distance: a new perspective on syntactic patterns in natural language'' by Haitao Liu et al}

\author{R. Ferrer-i-Cancho}

\ead{rferrericancho@cs.upc.edu}

\address{Complexity \& Quantitative Linguistics Lab, LARCA Research Group, Departament de Ci\`encies de la Computaci\'o, Universitat Polit\`ecnica de Catalunya, Campus Nord, Edifici Omega Jordi Girona Salgado 1-3. 08034 Barcelona, Catalonia (Spain)}

\end{frontmatter}

% +/- 6000 character

Liu et al's reflections on the term dependency length minimization \cite{Liu2017a} may look anecdotal but they are not. By the turn of the 20th century, we put forward a ``Euclidean distance minimization'' hypothesis for the distance between syntactically linked words and various word order phenomena \cite{Ferrer2004b,Ferrer2008e} \footnote{These were pieces of our PhD thesis \cite{Ferrer2003b} that were submitted for publication before its defense. }. 
Later on, pressure from language researchers forced us to replace it with terms such as ``online memory minimization'' \cite{Ferrer2013e} because our initial formulation was obscure to them. 
Recently, researchers from all over the world have been granted to use the term ``dependency length minimization'' by the popes thanks to whom \cite{Futrell2015a} came into light.
Although ``length'' is a particular case of distance in this context and thus downsizes our original formulation, it is still abstract enough to allow for progress in theoretical research \cite{Ferrer2014c} and frees us from the heavy burden of contingency, {\em i.e.} the real implementation of the principle (at present believed to result from decay and interference as reviewed by Liu et al) or the current view of the architecture of memory \cite{Jonides2008a,Lewis2006a}. Our position is grounded on the high predictive power of that principle {\em per se} \cite{Ferrer2013e}.

However, the lower generality of the term ``dependency length'' can be an obstacle to the construction of a fully-fledged scientific field \cite{Bunge1984a}. First, distance minimization allows one to unify pressure to reduce dependency lengths (still distances) with constraints on word order variation and change arising from a principle of swap distance minimization \cite{Ferrer2016c}. ``Distance minimization'' has therefore a higher predictive power and greater utility in a general theory of communication. Second, distance provides a ``formal background'' or a ``specific background'' (following Bunge's terminology \cite{Bunge1984a}) from physics or mathematics such as the theory of geographical or spatial networks (where the syntactic dependency structures of sentences are particular cases in one dimension) \cite{Reuven2010a_Chapter8,Gastner2006b} or the theory for the distance between like elements in sequences (where the couple of words involved in a syntactic dependency are particular cases of like elements) \cite{Zornig1984a}. Therefore we agree with \cite{Liu2017a} on the convenience of the term distance. 

A less flashy contribution of \cite{Futrell2015a} has been promoting the need of controlling for sentence length (as a predictor of dependency length in their mixed-effects regression model) in research on dependency length minimization, an important methodological issue \cite{Ferrer2013c} that was addressed early \cite{Ferrer2004b} but neglected in subsequent research ({\em e.g.}, \cite{Liu2008a,Temperley2008a,Gildea2010a}). % add Gildea2007a in case of free space ??? 
% Gildea, D., & Temperley, D. (2007). Optimizing grammars for minimum dependency length. In Proceedings of the Association for Computational Linguistics (pp. 184–191). Prague: Association for Computational Linguistics.

Liu et al focus their review on the fundamental principle of dependency length minimization but understanding how it interacts with other principles is vital. In 2009, we put forward another fundamental word order principle, {\em i.e.} predictability maximization, and presented a theoretical framework culminating in a conflict between dependency length minimization and predictability maximization \cite{Ferrer2009h}. For sociological reasons, these arguments started appearing in print many years later \cite{Ferrer2014a,Ferrer2013e,Ferrer2013f}. For the case of a single head and its dependents, the minimization of dependency lengths yields that the head should be placed at the center of the sequence whereas the principle of predictability maximization (or uncertainty minimization) yields that the head should be placed at one of the ends of the sequence (last if the head is the target of the prediction; first otherwise) \cite{Ferrer2013f,Ferrer2014a}. 

Liu et al review two major sources of evidence of dependency length minimization: the analysis of dependency treebanks and psychological experiments. 
A critical difference between them is that the former is based on the calculation of the total cost of the sentence (as a sum or mean of all the dependency lengths of the sentence) while the latter is based on a partial calculation and thus it can be misleading. Suppose that one wishes to compare the cost of two orderings of the same sentence. The observation that the processing cost of a sentence decreases when the length of a dependency increases, does not allow one to conclude that dependency length minimization cannot explain the results because shortening an edge implies moving at least one of the words defining it, and every move could imply the reduction of other edges eventually reducing the total sum of dependency lengths or altering the so-called complexity profile ({\em e.g.}, \cite{Morrill2010a}), rendering fair comparison impractical. The problem of partial calculation of length costs has already been discussed in the context of research on the cost of crossing dependencies \cite{Ferrer2016d} and worsens when the sentences being compared differ not only in order but also in content. Another challenge is the precision of dependency length that is typically measured in words. Lengths in phonemes or syllables shed light on why SVO languages show SOV order when the object is a short word such as a clitic \cite{Ferrer2014e}. 

Without addressing these issues, the anti-locality effects or long-distance dependencies reviewed by Liu et al can neither be attributed to predictability maximization nor be interpreted as a violation of dependency length minimization safely; an effective evaluation of the theoretical framework above can be impossible (as that framework makes theoretical predictions based on the calculation of full length costs).
The real challenge for psycholinguistic research is not the extent to which the theoretical framework above is supported by current results in the lab but rather to increase the precision of dependency length measurements and investigate the experimental conditions in which the following theoretical predictions are observed\cite{Ferrer2014a,Ferrer2013f}: one principle beating the other, coexistence, collaboration between principles or the very same trade-off causing the delusion that word order constraints have relaxed dramatically or even disappeared. This is the way of physics. 

Our concern for units of measurement is not a simple matter of precision but one of great theoretical importance: if the length of a dependency is measured in units of word length ({\em e.g.}, syllables or phonemes) then it follows that the length of a dependency will be strongly determined by the length of the words defining the dependency and that of the intermediate words. Therefore, pressure to reduce dependency lengths implies pressure for compression \cite{Ferrer2012d,Gustison2016a}, linking a principle of word order with a principle that operates (non-exclusively) on individual words. An understanding of how the principle of dependency length minimization interacts with other highly predictive principles beyond word order is a fundamental component of a general theory of animal behavior that has human language as a particular case.
    
\section*{Acknowledgments}

We thank C. G\'omez-Rodr\'iguez and H. Liu for helpful comments. This research was funded by the grants 2014SGR 890 (MACDA) from AGAUR (Generalitat de Catalunya) and also the APCOM project (TIN2014-57226-P) from MINECO (Ministerio de Economia y Competitividad).
       
\bibliographystyle{elsarticle-num} 
% \bibliography{../biblio/philosophy,../biblio/complex,../biblio/rferrericancho,../biblio/cl,../biblio/cs,../biblio/ling,../biblio/maths}

\end{document}